# Vegetation Mapping by UAV Visible Imagery and Machine Learning

G. Vitali
University of Bologna - Italy
giuliano.vitali@unibo.it

**Abstract**
An experimental field cropped with sugar-beet with a wide spreading of weeds has been used to test vegetation identification from drone visible imagery. Expert masked and hue-filtered pictures have been used to train several Machine Learning algorithms to develop a semi-automatic methodology for identification and mapping species at high resolution. Results show that 5m altitude allows for obtaining maps with an identification efficiency of more than 90%. Such a method can be easily integrated to present VRHA, as much as tools to obtain detailed maps of vegetation.

## 1. Introduction

The analysis of Earth surfaces has been pursued for years for a number of scopes, including biodiversity mapping, cropland survey. These tasks have been for a long time based on remote imaging (satellite, aircraft) (e.g., Xie et al., 2008; Zaitunhah et al., 2021). Though satellite analysis made huge steps forward in surface survey (e.g., Boiarskii, 2020, d'Andrimont et al., 2021), aircraft-based scanning missions gave an important supplementary aid in collecting more detailed information, which in some cases can be automated using UAVs (e.g., Zhang et al., 2019). While satellites and aircrafts may only deliver images with a long time interval with respect to cover dynamics, such autonomous vehicles may strongly reduce the interval between images. In UAVs two moreadded values can be identified - as a cheap technology, they are affordable to many people, and they may fly close to the surface allowing to get images with the desired resolution.
UAV picture taken from several flight hight are have been used to map invasive vegetation in desertic areas (Kedia et al., 2021) or species identification in shrubland (Jimene and Diaz-Delgado, 2015), though a far larger interest is focused on weed control (Guan et al., 2015, Sanchez-Sastre et al.,2020).
Some weeding machineries for Variable Rate Herbicide Application (VRHA) are already equipped with ground-based cameras whose pictures are used to detect weed - recognition though the low efficiency is witnessed by search of integration with a-priori knowledge (planting rows), and limited targeting to weeding in the early stage of development (Wang et al., 2019).
Vegetation canopy is a dynamically structured biological system made of species strongly interacting to one another, and recognition of species at different development stages and within a community depends on several factors. Distance at which we observe a landscape and the season, as much as cloudiness, daytime, and angle of view may significantly affect a number of features making images considerably different to one another (El Faki et al., 2000). When looking up close enough, 3D-cameras can help perceive elements at several layers (Gai et al., 2018; Papp, 2021), but in most observations the picture appears just as a flat one, and reflectance and shading represent a source of noise, hardly to be removed. To solve these issues, two are the main approach, one focusing on sensors, and another on colour space models.

Former RGB CCD-based camera approaches date back more than 20 years (e.g. Perez et al., 1997) and limitations of RGB colour space is wel known (e.g., Bai et al., 2013) so that several alternative colour models have been tested (e.g., Hernández-Hernández et al., 2016).

Multi-spectral cameras (Dubbini et al, 2017), well known in satellites approach (e.g. Jimenez and Diaz-Delgado,2015), and vegetation oriented 'true-colour' sensors (Schmittman and Lammers,2017) are some of the ways used to obtain spectrum-related features, together with hyper-spectral and UV ones, the latter aimed at insulating reflectance or induced fluorescence (Liu and Bruch, 2020). A NDVI-based approach well known in satellite applications already proved effective in detecting early spreads of Sinapis arvensis from UAV images in row crops (Sanchez-Sastre et al., 2020).

Alternative colour models have been used (Forero and Manero-Rivera, 2019; Fitriyah and Wihandika, 2018; Schmittmann and Lammers, 2017), including HSV which interprets human colour perception system by means of a hue value together with a components related to enlightenment and shading (value) and to surface richness of pigments (saturation). Hue colour space mode has already been used for image segmentation (e.g. Mao et al., 2008).

Together with colour other image features can be extracted from an image, some with a regular component (as patterns), others noisier, called texture (e.g. Yang et al., 2021). While patterns just scale with distance, texture strongly depends on resolution, which in the case of drones depends on flight height. A series of statistical texture features are described by Armi and Fekri-Ershad (2019) offering evidence of their use.

Textural features can be approached with a structural, statistical and time series approach. Structural analysis could reveal the existence of patterns, namely repeated/regular regions called texels/textons. As getting closer to the subject (e.g., a species individual) information could increase considerably but self-similar/homothetic patterns only put in evidence when reaching the scale of characteristic dimensions and not for every species (see e.g., figure 1).

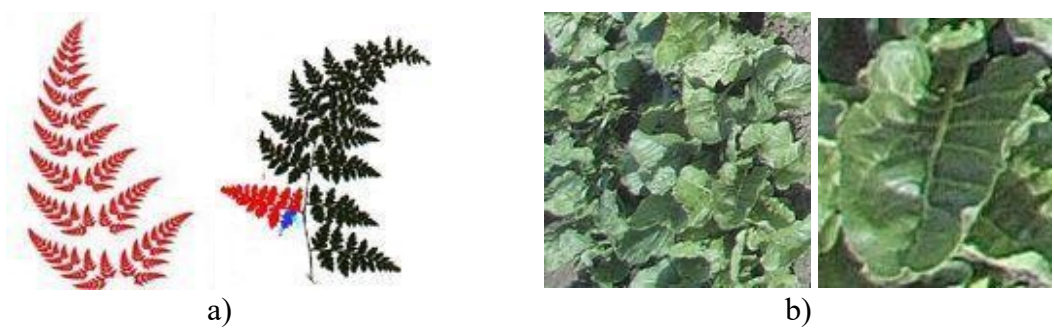

a) b)

Figure 1. Drawing of fern leaf (a) characterised by leaf self-similarity, and images of leaves of a sugar-beet taken at different distances (b).

At a lower scale, several Machine Learning (ML) techniques have been adopted for species identification (Wäldchen and Mäder, 2018). Solutions range from single leaves when focusing on particular species (e.g. Wu et al., 2022; Xu et al., 2022) or from close-up picture taken from autonomous weeding vehicles (e.g. Joshi, 2017; Potena et al., 2016; Su et al.,2021). Convolutional Neural Networks (CNN) already proved to be useful in driving autonomous vehicles in the middle of a grown crop (e.g. Abdullahi et al., 2017). As a major bottleneck of assisted ML is represented by a collection of training pictures, the participative approach has proved to be very effective (Affouard et al., 2017). To the scope, a technique gained a high popularity in image classification - deep Neural Networks (NN) can be pre-trained on certain data sets to obtain 'embedders' to be used to extract features from training images. Such pre-learned embeddings can be used to significatively reduce the learning time (see e.g., Rawat and Wang, 2017). A last option is represented by non-assisted techniques. Integration of colorspace features to non-assisted tools (e.g. Self Organised Maps-SOM and Improved Particle Swarm Optimum-IPSO) seems to give fruitful results for VRHA (Xu et al., 2018), also in mulched soil (Golzarian,2011), though they seem not suitable when multiple species are to be recognised .

The majority of the solutions described so far try to simplify recognition focusing on a single target species (the crop), or seedlings in the early stage of a crop or, on the contrary, looking at the landscape from a distance. To the scope use close up images to capture the detail of a single species are used, or observing the earth surface from a scale allowing to make inhomogeneities negligible. The scope of this work is to analyse vegetation from an intermediate distance/scale, useful both for a detailed vegetation mapping as much as for late weeding. As the supervised approach todate seems the more precise, the need of supplying a large enough set of training pictures and reduction of training time are also pursued.

**The present study aims** at developing and testing a methodology to map vegetal species from UAV images, based on a semi-automated method for collecting training images. Different learners are further compared and a validation is performed, aimed at estimating the efficacy of the approach. Results are further presented and discussed, and final conclusions are drawn to describe the range of possible applications.

## 2. Materials and Methods

The proposed method is based on the following steps:
1. site identification - finding a site suitable for species recognition and allowed to UAV flights;
2. shooting - planning missions, performing flights, extracting pictures taken by UAV and preliminary image analysis with the scope to identify macro-areas based on the observation of density of vegetal species (crops or weeds);
3. preprocessing - conversion of images to the hue-colour model and selection of images for training of ML
4. training and classification, based on extraction of features, ranking, selection of classifiers, training and test;
5. validation on different pictures.

Additional methodological details are described in the following sections.

### 2.1. Site identification

The site selected for the survey was an experimental field (c.ca *300m* long and 50m wide) located in North Italy (near Bologna, lat. 44°33', long. 11°25'). The field is cropped with rain-fed and untreated sugar-beet (i.e., Beta *vulgaris* var. saccharifera, Bv). In the period of observation (advanced vegetative stage) it showed large patches of Sinapis *arvensis (Sa),* Chenopodium *album (Ca),* two typical weeds of Mediterranean area (Figure 2), and bare soil (S).

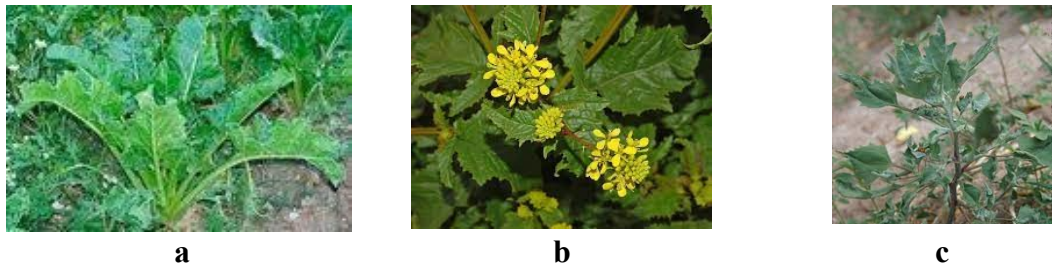

|   a   |   b   |   c   |

Figure 2. Plants under examination: a) Beta *vulgaris*; b) Sinapis *arvensis (photo by wikipedia)*; c) Chenopodium *album (photo by swbiodiversity.org)*.

The field represents an extreme case of a crop with an anomalous spreading of weeds, hardly representing a real case, which may also be representative of an ecological transition of an abandoned field moving to self-renaturalization. Therefore, the site may be seen both, as a ***simplified landscape populated by few herbaceous species*** and as a ***field crop in a late weeding context***.

## 2.2 UAV mission

A commercial esa-copter (Typhoon H, by Yuneec,2020), equipped with a 4K resolution rSGB camera (model CGO3+3.2.34, f/2.8, exp time 1/15, sens. ISO-100, FOV: 98°, image width:4096, heigh:2160, depth:24), was used to perform mission flights on the field at different altitudes: 5, 7, 15 and 35m.

Flights took place in late spring (May 8, 2017 - 2.00-5.00PM, sun elevation c.ca 35°), and pictures have been taken pointing the camera downward ('flat lay') with a regular time interval set on the base of height and velocity (to ensure superposition of images of about 20%). Fig.3 shows a sample of images taken at *35m height* where large patches of S. *arvensis* in blossom (yellowish area) and C. *album* (bluish on middle top) in the sugar-beet field are well visible, together with patches of bare soil.

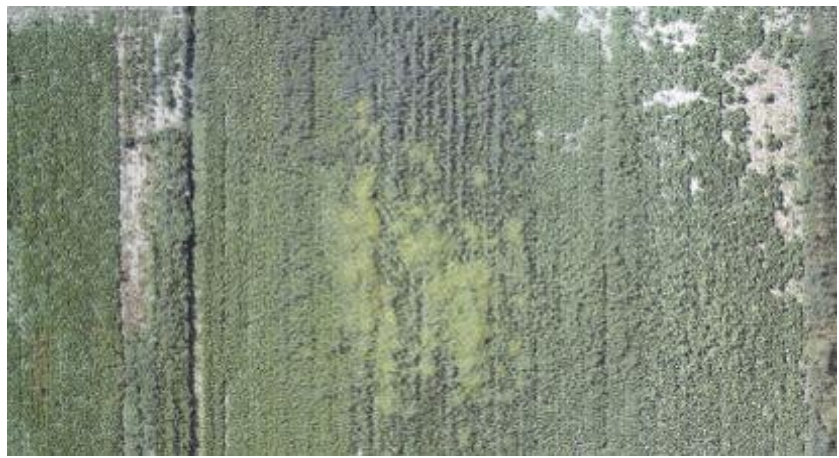

Figure 3. Aerial image of the untreated experimental sugar beet field at *35m* height.

## 2.3 Image selection for ML training

After considering the available collection together with experts (botanists and farmers), pictures taken at *5m* height have been selected for the study (resolution of *1px < 1cm*). Among them two pictures have been used to select the 4 more relevant types of cover, Fig.4a for S.arvensis and Bare Soil, and Fig4b for C.album and B.vulgaris.

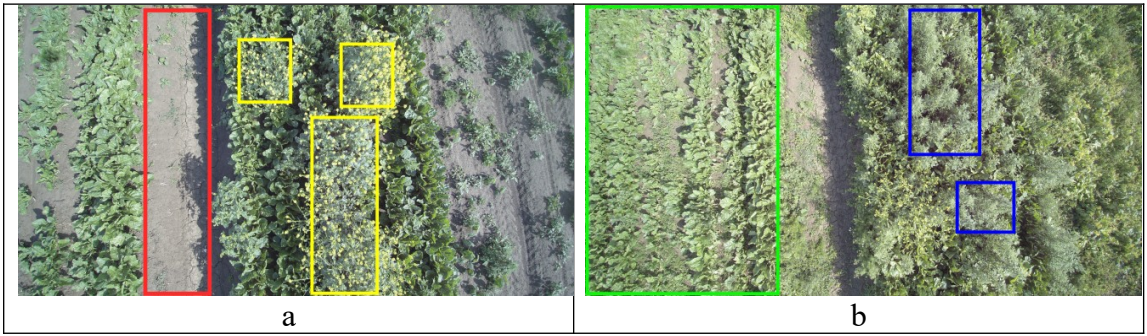

Figure 4. Pictures selected for identification of relevant type of cover. Rectangles of different colours are used to show the 4 covers used for this study: red-S, yellow-Sa, green-Bv, blue-Ca.

Pictures have been successively split into tiles, from which a selection has been taken for the successive ML training process. Three tile sizes have been considered: *64x64px* (*c.ca 25 cm²*), *124x124px* (*c.ca 100 cm²*), 256x256px (*c.ca 400 cm²*), as shown in Fig. 5.

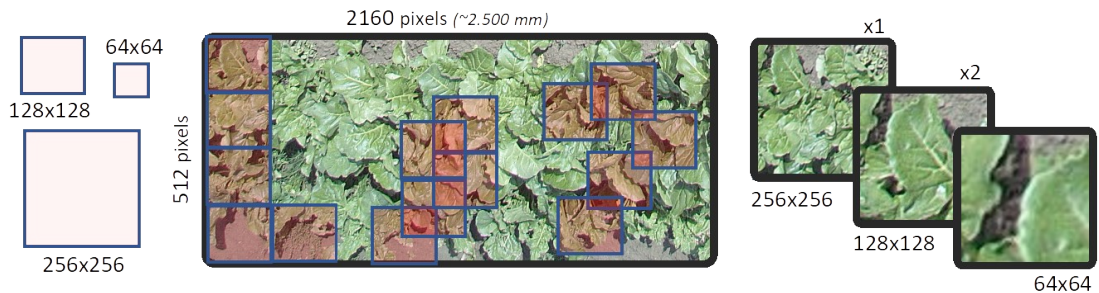

Figure 5. Representation of tiles of different size

**Selection criteria** - Two criteria have been adopted to select tiles for ML training, the *direct spotting of tiles* from the images, and two *semi-automatic selection strategies*.
As told before the more efficient methods of recognition are the human-assisted ML techniques, meaning that experts should support images to be used for training the recognition system.
In the first method an expert has been asked to identify and categorise tiles *256x256px*.
Two semi-automatic methods have been issued to increase the number of training tiles. The first starts from clustering of images of the previous first set, aimed at suggesting a set of tiles more similar to the optimal ones, reducing the risk to select tiles too far in their contents.
The second semi-automatic method is based on masks, each derived from a different coverage, also drawn by an expert. The masks have been used to select a larger number of tiles of lower size. Selection is based on two criteria, mask overlay threshold (TSH) and hue value. To the scope hue spectra are derived from masked regions and used to define a range.

## 2.4   Machine learning

ML analysis has been performed using Orange (2021, ver. 3.30), an open-source ML platform providing a workbench allowing the visual building of workflows for data classification, clustering, cross-validation, and prediction, that already proved to be effective for image analysis (Godec, 2019).
The workflow developed for the present analysis includes the following steps.

- targeting category: assign training image collection (by folders' name), being their categorization already including hue range information;
- image embedding: images are passed to an embedder to build a feature matrix made of an enhanced data table with image descriptors. Orange includes the following embedders: '*SqueezeNet*', '*Inception-v3*', '*VGG-16*', '*VGG-19*', trained on ImageNet (2021), '*DeepLoc*' trained on yeast cell images, and '*Painters*', based on paintings (details and references in Orange, 2021);
- ranking: selection of most relevant features;
- cluster analysis: based on embeddings, it is aimed at a preliminary assessment of image clustering, and to identify borderline cases. The 'cosine' distance has been chosen because of its proven ability to minimise distortions related to the scale. Learners' selection: Orange includes 18 different learning models - as there is no a priori method to determine the best one(s) with respect to a specific dataset, a cross-validation procedure has been adopted, with stratified folding technique, adopting an automated random tile extraction.
- Validation and predictions. A final assessment of methodology has been performed in two different ways. For direct tile selection validation has been carried out by randomly sampling tiles (10%) from the original training dataset, but, for each test, the dataset has been stripped only of the tile under examination, so as to leave the dataset almost unchanged. In the semi-automatic procedure, the validation has been made on a different picture from the field (also from *5m* height) and predictions are shown to an expert.

## 3. Results and discussion

**3.1 Direct selection** - Fig.4a and 4b have been split in *256x256px* tiles from which a direct selection been on allowed detecting B. *vulgaris* in 23 tiles, S. *arvensis* in 56, C. *album* in 26 and the bare soil was in the other 35. Selected tiles can be seen in Figures 6,8 and 9.

The embedders included in Orange have been tried preliminarily, and the best performances have been shown by '*Painters*', an embedder obtained by a Convolutional Neural Network trained on 79.433 images of painting by 1.584 painters with the scope to predict painters from artwork images (Ilenic, 2021). The 140 training images (tiles) were converted into a data table of 2048 features.

As a preliminary analysis, a Cluster Analysis (CA) has been performed. With a max depth of 10, the 23 tiles of B.*vulgaris* were always grouped together. In a sample of images reported in Fig.6, it can be seen that the closest tiles from the reference one (for B.vulgaris), show minimal differences in shape and thickness of leaves. Moving to the right, elements of other categories enter the graph, starting from C. *album*, then followed by bare soil and C.*arvensis* at last, which is clearly different in terms of shapes, sizes, and colours.

Such a preliminary clustering has been repeated on other image samples, and all of them showing no discrepancies in clustering of training data, therefore we could proceed to the following step.

Classification has been performed using six of the most common classification models out the 18 available on Orange: Classification Tree (CT), k-Nearest Neighbours (kNN), Logistic Regression (LR), Neural Network (NN), Support-Vector Machines (SVM) and Random Forest (RF). Although it would be easy to include the other models in the analysis, any enhancement was evident from a preliminary screening.

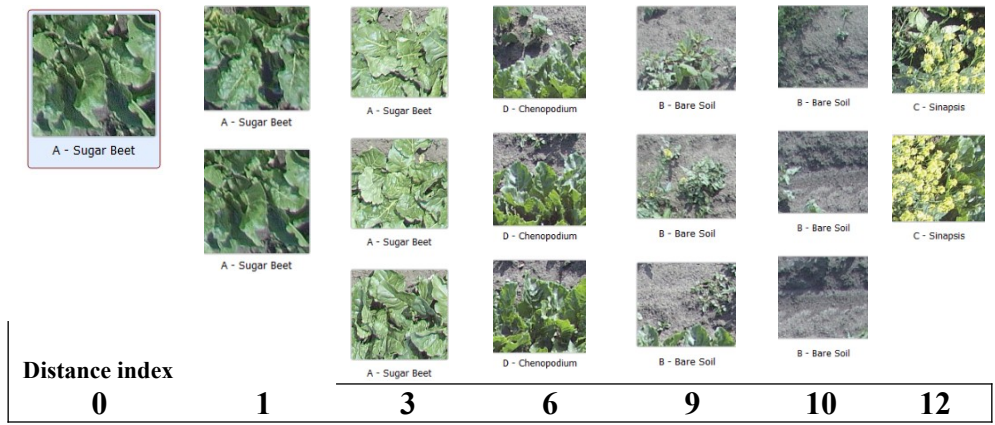

Figure 6. Sample of tiles together with their progressive distance

The AUC performance indicator (Area Under the receiver operating characteristic-ROC Curve), scored 0.991 for LR, 0.983 for kNN, 0.976 for SVM and slightly less for others, (the worst being CT with an accuracy of 0.796). Other evaluation criteria have been tried (Classification Accuracy, F1, Recall, Logloss and Specificity) without relevant differences in scoring.

Confusion matrices have been then produced to verify the agreement of classifiers, to estimate the risk for a wrong prediction. In table 1 three confusion matrices are shown for the most performing classifiers, NN and RF, together with CT, the less performing one.

|  |  | Neural Networks | | | | | Random Forest | | | | | Classification Tree | | | |
|---|---|---|---|---|---|---|---|---|---|---|---|---|---|---|---|
|  |  | *Bv* | *Bs* | *Ca* | *Sa* |  | *Bv* | *Bs* | *Ca* | *Sa* |  | *Bv* | *Bs* | *Ca* | *Sa* |
| B.vulgaris | Bs | 100.0 | | | | Bs | 100.0 | | | | Bs | 68.2 | 9.1 | 2.1 | 4.2 |
| bare soil | Bs | | 93.9 | | | Bs | | 93.9 | | | Bs | 22.7 | 75.8 | 2.1 | |
| C.album | Ca | | 3.0 | 88.5 | 23.8 | Ca | | 6.1 | 77.4 | 16.7 | Ca | | 12.1 | 76.6 | 50.0 |
| S.arvensis | Sa | | 3.0 | 11.5 | 76.2 | Sa | | | 21.0 | 81.3 | Sa | 9.1 | 10.0 | 19.1 | 45.8 |

Table 1 -. Confusion matrices with the probability of errors in prediction for three selected learners.

It is evident, e.g., that NN and RF never make confusion in the case of sugar-beet correctly predicting it in 100% of cases, while CT is right only at 68.2%.
Bare soil is also well recognized by NN and RF (93.9% of accuracy) while the remaining cases (6%) are considered synapsis by RF and not distinguished by NN. Furthermore, NN and RF show a certain difficulty (between 11.5% and 23.8%) in recognizing C.*album*, while NN seems weaker on S.*arvensis*.

**Validation** was carried out by randomly extracting a sample of 14 tiles from the dataset to be used for testing (*10%* of 140), 3 for B.*vulgaris*, 4 for bare soil, 4 for S.*arvensis*, and 4 for C.*album*. A distance graph has been used to check the randomness of sampling (Fig.7).

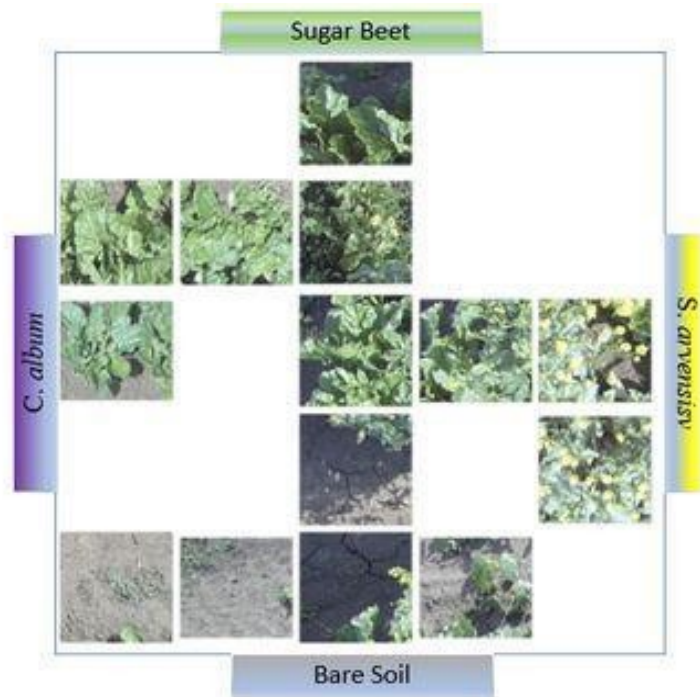

Figure 7. Images grid highlighting image classification

A detail of predictions made from the 6 classifiers on 4 randomly selected tiles are available in Table 2, where it is reported the actual and predicted categorization in terms of probability. For instance, in the case of the first row, the LR classifier correctly foresees B.*vulgaris* and bare soil both with a probability of 99%. Then, it suggests considering the 3$^{rd}$ image, initially clustered as S.*arvensis,* as soil since the probability of correspondence with S.*arvensis* is around 29% (with a probability of 71%). Similarly, it suggests considering the 4$^{th}$ image as S.*arvensis* instead of C.*album* even if in this case the difference in probability is marginal (53% vs 47%).

| Learner | B. *vulgaris* | Bare soil | S. *arvensis* | C. *album* |
|---|---|---|---|---|
| LR | 0.99 | 0.99 | 0.29 (soil) | 0.44 (S. *arvensis*) |
| NN | 1.00 | 1.00 | 0.00 (soil) | 0.94 |
| RF | 0.90 | 0.80 | 0.00 (soil) | 0.41 |
| TR | 1.00 | 1.00 | 0.00 (soil) | 0.00 (B. *vulgaris*) |
| SVM | 0.95 | 0.78 | 0.40 (soil) | 0.67 |
| kNN | 1.00 | 1.00 | 0.20 (soil) | 0.60 |

Table 2. Classification proposed by used classifiers, expressed as the probability to confirm the initial clustering.

Observing these and the other results it becomes evident how the learners work are very precise. In fact learners, together with having recognized the B.vulgaris with a high accuracy (*78% -100%)*, were also able to recognise soil both under the sunlight and in the shadow and contaminated by S.arvensis. They also found that the 3rd tile in Table 2, labelled as S.*arvensis*, should be better labelled as bare soil. All classifiers recognize a potential misclassification (of which we are now certain), indicated by 60% to 100%. The 4th photo is the most uncertain. This depends on the contents of the photo itself where the presence of C. album is observed against a background of B.*vulgaris*.

**3.2. Semi-automatic tile selection I - embedding and cosine clustering** - The embedding procedure can be used also without a preliminary categorization of tiles, and hierarchical clustering can be obtained on the base of cosine distance.
It is evident from Fig 8 where two different clusters are displayed by way of example.

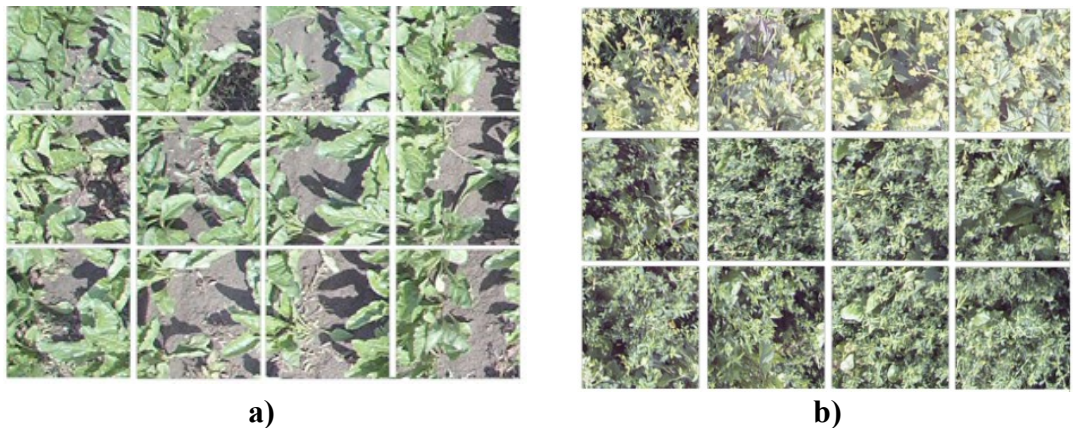

a) b)

Fig 8. Two independent groups of images clustered by distance metrics.

In Fig 8b it is also evident how S. *arvensis* and C. *album* are categorised in the same macro-cluster with this method.
The distance metrics can be conveniently applied with the scope of searching for neighbours,to be used as a helper to identify images to be collected for the training dataset.

**Validation** was performed for this method on a focus species, the crop one, using the 6 classifiers taken into consideration for this analysis. NN allowed the recognise B. *vulgaris* with a probability over >50% on 51 tiles out of 128. The other classifiers provided similar predictions but reporting smaller groups: kNN, 34 images (with >40% probability); LR 27, (>51%); RF, 27 (>30%); SVM, 22. However, these image datasets were superimposed for the most part with NN and kNN methods able to fully identify all the independent elements: 57 tiles on 128, equivalent to 44.5% of the cultivation field. Results were manually checked confirming the correct attribution. Some identifications are reported in Fig. 10, together with their prediction probability (by NN).

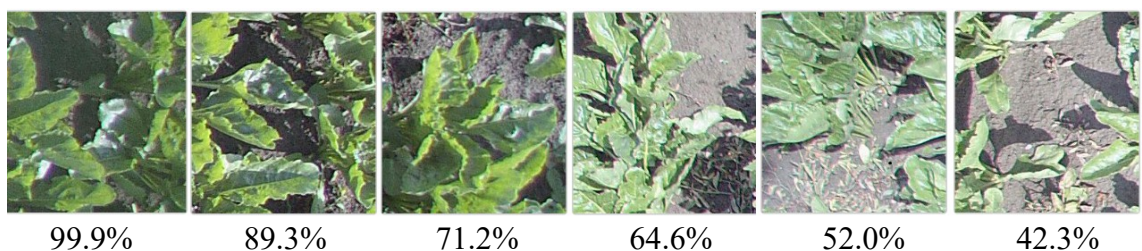

99.9%      89.3%      71.2%      64.6%      52.0%      42.3%

Fig 9. Predictions made by machine learning (i.e., Neural Network): the foliage identification (B. *vulgaris*) is proposed by related probabilities.

**3.3 Semi-automatic tile selection II - painting and hue filtering** - From the painting process, 4 masks have been prepared which are shown in Figure 10.

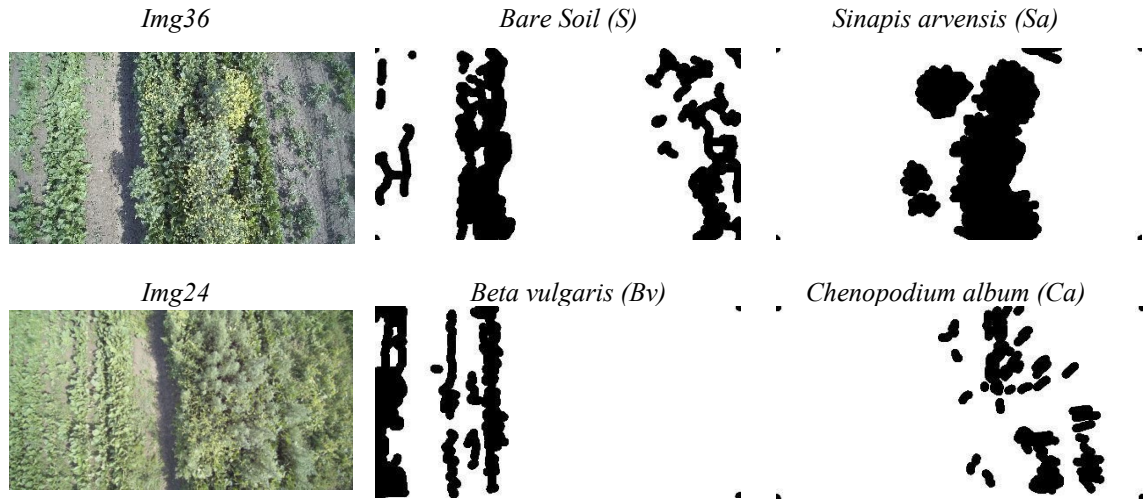

Figure 10. Picture selected for identification of relevant type of cover, and masks use for obtaining images for training ML.

Masks have been further filtered using hue ranges. To the scope the RGB pictures have masked and successively converted to the HSV colour space (algorithms can be found e.g., in Agoston, 2016). Finally their hue spectra have been used to identify the hue range (in the scale 0-360) of each cover (Fig.11). Ranges adopted have been [55-125] for B. *vulgaris* and C. *album*, [55-165] for S. *Album* and two ranges [25-55; 210-235]. The wider spectrum of S. arvensis was due to its flowers, whereas soil shows the effect due to albedo and shadowed areas.

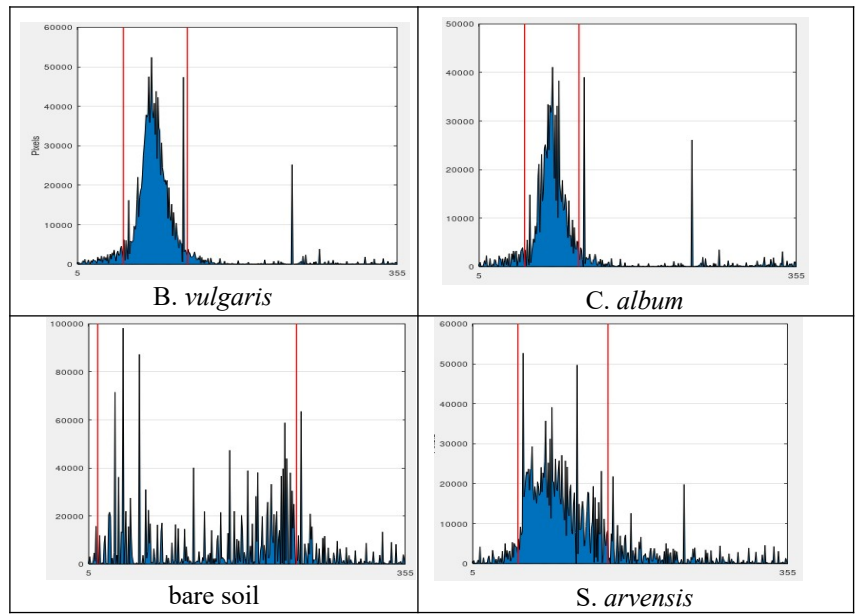

Figure 11. Visualisation of hue spectrums of the 4 masked coverages.

Tile selection produced samples with a variable numerosity depending both on tile size and STH (see Table 3). Each sample has been obtained using 3 shifts (as a tile multiplication factor).

The more uniform coverage (soil) is the one yielding a higher level of tiles, followed by the main crop (B. *vulgaris*) while it was more difficult to harvest a sufficiently high number of tiles from wild species, because of the effect of size and the superposition constraint (STH) is relevant.

| Pixel | STH | B. *vulgaris* | C. *album* | Bare soil | S. *arvensis* |
|---|---|---|---|---|---|
| 64x64 | *0.98* | 398 | 207 | 510 | 719 |
| 64x64 | *0.99* | 379 | 189 | 419 | 664 |
| 64x64 | *0.999* | 257 | 115 | 170 | 315 |
| 128x128 | *0.8* | 90 | 57 | 140 | 177 |
| **128x128** | ***0.9*** | *62* | *39* | *89* | *158* |
| 128x128 | *0.95* | *47* | 28 | 74 | 145 |
| 256x256 | *0.5* | 34 | 21 | 42 | 46 |
| 256x256 | *0.6* | 20 | 15 | 29 | 41 |
| 256x256 | *0.7* | 15 | 8 | 19 | 35 |

Table 3. Tile set consistency for different tile size and level of superposition (STH)

The 9 data-sets gave good cross-validation values (3 fold-layering) with all the 6 classifiers, with a value always greater than 90%. It can be observed that the number of images doesn't affect the precision of more precise learners while slightly increases the precision of less precise ones (see Appendix 1). Such a steadiness of values is a sign of high robustness of the method and a low risk to adopt a wrong learner. In sum, a data-set of at most 400 tiles seems to be sufficient to have an efficient training together with a contained processing time - kNN and Random Forest are the best ones from this viewpoint.

Validation has been at last performed on a different picture (Fig. 12) which has been splitted in 496 (*16x31*) *128x128px* tiles.

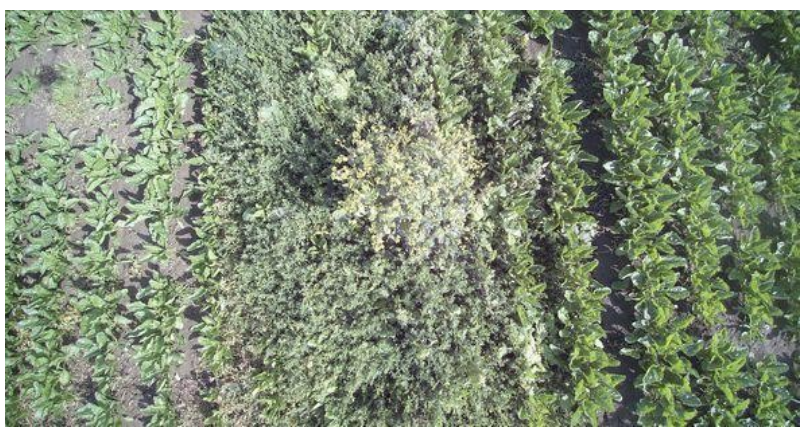

Figure 12 - Picture at 5m height taken from the experimental field used for method validation.

Prediction, performed using the learners trained before, have been submitted to experts, who selected as the better identification the one performed by NN learner. Tiles ascribed to B.vulgaris and S.arvensis are shown in Fig.13.

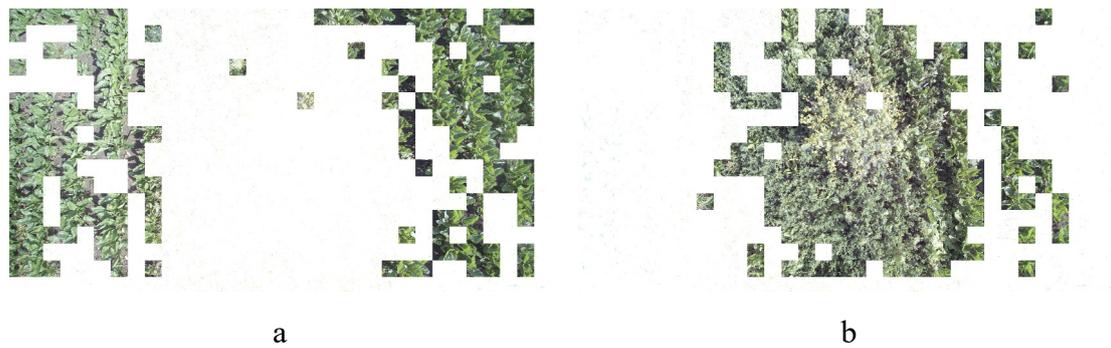

a　　　　　　　　　　　　　　　b

Figure 13 - Images showing the tiles ascribed from NN learner to B.vulgaris (a) and C.album (b).

**4. Conclusions**

The present article aims at identifying and testing a quick and simple methodology which combines UAV (drone) survey missions, rapid and incremental manual recognition of species based on samples of aerial pictures, and automated generation of training images for ML algorithms.

Images from 5m height taken from a flying drone can be very suggestive in interpretation of herbaceous vegetation and able to supply pictures helpful in automatic image segmentation to be used for mapping, suitable to be applied in wild area surveys or Variable Rate Herbicide Application. The methodology, based on human-assisted technique, responds well both in the case of a direct selection of tiles to be used for training a learning system, or in the case the tiles are selected automatically from painted masks. In both cases prevision forecasts are largely ascribed to tiles which havebeen assigned a bad label, preventing the training procedure to learn correctly.

The semi-automatic generation of tiling for training purposes integrating expert painting and hue filtering of pictures from a common digital camera has also shown effectiveness.

Therefore we can conclude that the lack of precision in the present case recognition does not depend on the learners, but instead from context difficult to be interpreted even from an expert, and leading to an erroneous categorization.

Also the goodness of results may strongly increase if there is the possibility to focus on a target species. In this case, B. *vulgaris* represents the crop grown, and therefore it is easily meant as the focus species for a (lately) weeding. In this case, all five classifiers (excluding CT) offer a *100%* accuracy,therefore we can consider the procedure as a good tool in High Resolution VRHA.

The resulting methodology is based on open-source algorithms, which profit from embedding techniques to reduce computational power to make the procedure portable to mobile devices and off-line contexts. The tool can be easily integrated to generate an image partitioned into categorised tiles to be used both in raster or vector GIS.

**Acknowledgments**
Author thanks Coprob and G. Campagna for experimental field availability, C.Andreasen, and F.Zingaretti for their hints on agronomic practises and spraying technology, F.Memmi for UAV driving, M.Ferrara, E.Magnanini and C.Fragassa for image processing discussions.
**Funding** - This research was funded by WeLaser H2020 project Grant Agreement N. 101000256. https://cordis.europa.eu/project/id/101000256.**Declaration of interests**
The author declare that they have no known competing financial interests or personal relationships that could have appeared to influence the work reported in this paper.

**Appendix 1 - Cross validation of semi-automatic data-sets**

| Tile size | STH % | Images | Model | Train time | Test time | AUC | CA | F1 | Precision | Recall | Logloss | Specificity |
|---|---|---|---|---|---|---|---|---|---|---|---|---|
| | | | | | | Cross-validation (n. of folds = 3, stratified) | | | | | | |
| 256 | 50 | 59 | kNN | 1.107 | 0.547 | **0.994** | 0.949 | 0.948 | 0.955 | 0.949 | 0.112 | 0.996 |
| | | | Logistic Regression | 2.291 | 0.636 | **1.000** | 0.983 | 0.983 | 0.986 | 0.983 | 0.044 | 0.999 |
| | | | Neural Network | 5.246 | 1.245 | **0.999** | 0.983 | 0.983 | 0.986 | 0.983 | 0.049 | 0.999 |
| | | | Random Forest | 1.117 | 0.623 | **0.990** | 0.949 | 0.935 | 0.953 | 0.949 | 0.315 | 0.874 |
| | | | SVM | 2.805 | 1.224 | **0.982** | 0.898 | 0.866 | 0.843 | 0.898 | 0.345 | 0.749 |
| | | | Tree | 0.863 | 0.000 | 0.829 | 0.864 | 0.848 | 0.832 | 0.864 | 4.683 | 0.905 |
| 256 | 40 | 160 | kNN | 1.486 | 0.605 | **1.000** | 0.990 | 0.990 | 0.991 | 0.990 | 0.036 | 0.998 |
| | | | Logistic Regression | 3.380 | 0.853 | **1.000** | 1.000 | 1.000 | 1.000 | 1.000 | 0.017 | 1.000 |
| | | | Neural Network | 5.832 | 1.609 | **1.000** | 1.000 | 1.000 | 1.000 | 1.000 | 0.001 | 1.000 |
| | | | Random Forest | 1.420 | 0.570 | **0.999** | 0.990 | 0.990 | 0.991 | 0.990 | 0.229 | 0.998 |
| | | | SVM | 3.846 | 1.395 | **1.000** | 0.990 | 0.990 | 0.990 | 0.990 | 0.145 | 0.989 |
| | | | Tree | 1.453 | 0.000 | 0.930 | 0.911 | 0.910 | 0.912 | 0.911 | 3.082 | 0.965 |
| 256 | 30 | 167 | kNN | 0.974 | 0.612 | **0.994** | 0.949 | 0.948 | 0.955 | 0.949 | 0.112 | 0.996 |
| | | | Logistic Regression | 2.459 | 0.853 | **1.000** | 0.983 | 0.983 | 0.986 | 0.983 | 0.044 | 0.999 |
| | | | Neural Network | 5.142 | 1.558 | **0.999** | 0.983 | 0.983 | 0.986 | 0.983 | 0.049 | 0.999 |
| | | | Random Forest | 1.041 | 0.785 | **0.981** | 0.881 | 0.857 | 0.834 | 0.881 | 0.344 | 0.868 |
| | | | SVM | 3.154 | 1.816 | **0.943** | 0.898 | 0.866 | 0.843 | 0.898 | 0.444 | 0.749 |
| | | | Tree | 0.964 | 0.002 | 0.829 | 0.864 | 0.848 | 0.832 | 0.864 | 4.683 | 0.905 |
| 128 | 60 | 308 | kNN | 1.658 | 1.140 | **1.000** | 0.981 | 0.980 | 0.982 | 0.981 | 0.046 | 0.994 |
| | | | Logistic Regression | 9.940 | 0.683 | **1.000** | 1.000 | 1.000 | 1.000 | 1.000 | 0.010 | 1.000 |
| | | | Neural Network | 11.897 | 1.733 | **1.000** | 0.994 | 0.994 | 0.994 | 0.994 | 0.022 | 0.998 |
| | | | Random Forest | 1.821 | 1.061 | **0.999** | 0.984 | 0.983 | 0.985 | 0.984 | 0.122 | 0.990 |
| | | | SVM | 5.862 | 2.061 | **1.000** | 0.997 | 0.997 | 0.997 | 0.997 | 0.058 | 0.994 |
| | | | Tree | 2.405 | 0.000 | **0.934** | 0.935 | 0.934 | 0.938 | 0.935 | 2.243 | 0.973 |
| 128 | 50 | 436 | kNN | 1.692 | 2.043 | **0.995** | 0.979 | 0.979 | 0.980 | 0.979 | 0.201 | 0.993 |
| | | | Logistic Regression | 17.064 | 1.683 | **1.000** | 0.993 | 0.993 | 0.993 | 0.993 | 0.020 | 0.998 |
| | | | Neural Network | 14.193 | 3.117 | **1.000** | 0.993 | 0.993 | 0.993 | 0.993 | 0.025 | 0.998 |
| | | | Random Forest | 1.994 | 1.014 | **0.998** | 0.972 | 0.972 | 0.974 | 0.972 | 0.153 | 0.991 |
| | | | SVM | 8.508 | 3.255 | **1.000** | 0.986 | 0.986 | 0.987 | 0.986 | 0.064 | 0.996 |
| | | | Tree | 5.362 | 0.006 | **0.989** | 0.986 | 0.986 | 0.986 | 0.986 | 0.408 | 0.994 |
| 128 | 40 | 580 | kNN | 2.426 | 2.761 | **0.998** | 0.976 | 0.976 | 0.977 | 0.976 | 0.119 | 0.992 |
| | | | Logistic Regression | 20.524 | 1.010 | **1.000** | 0.993 | 0.993 | 0.993 | 0.993 | 0.031 | 0.998 |
| | | | Neural Network | 16.577 | 2.689 | **0.999** | 0.983 | 0.983 | 0.983 | 0.983 | 0.059 | 0.994 |
| | | | Random Forest | 2.403 | 1.022 | **0.995** | 0.955 | 0.955 | 0.956 | 0.955 | 0.209 | 0.988 |
| | | | SVM | 9.759 | 2.623 | **1.000** | 0.984 | 0.984 | 0.985 | 0.984 | 0.060 | 0.995 |
| | | | Tree | 7.009 | 0.007 | **0.938** | 0.917 | 0.917 | 0.917 | 0.917 | 2.077 | 0.974 |
| 64 | 50 | 1969 | kNN | 7.420 | 18.023 | **0.995** | 0.974 | 0.974 | 0.975 | 0.974 | 0.233 | 0.992 |
| | | | Logistic Regression | 166.631 | 2.341 | **0.999** | 0.988 | 0.988 | 0.988 | 0.988 | 0.040 | 0.997 |
| | | | Neural Network | 99.580 | 3.896 | **0.999** | 0.988 | 0.988 | 0.988 | 0.988 | 0.050 | 0.997 |
| | | | Random Forest | 7.381 | 2.605 | **0.995** | 0.964 | 0.964 | 0.965 | 0.964 | 0.213 | 0.990 |
| | | | SVM | 37.451 | 7.542 | **0.998** | 0.988 | 0.988 | 0.988 | 0.988 | 0.050 | 0.997 |
| | | | Tree | 40.621 | 0.003 | **0.940** | 0.918 | 0.917 | 0.917 | 0.918 | 1.620 | 0.973 |
| 64 | 60 | 1555 | kNN | 4.538 | 11.534 | **0.997** | 0.981 | 0.981 | 0.982 | 0.981 | 0.151 | 0.995 |
| | | | Logistic Regression | 71.702 | 1.195 | **0.999** | 0.990 | 0.990 | 0.990 | 0.990 | 0.034 | 0.998 |
| | | | Neural Network | 69.050 | 3.222 | **0.999** | 0.992 | 0.992 | 0.992 | 0.992 | 0.046 | 0.998 |
| | | | Random Forest | 4.897 | 1.217 | **0.996** | 0.976 | 0.975 | 0.977 | 0.976 | 0.177 | 0.994 |
| | | | SVM | 25.244 | 4.762 | **0.999** | 0.989 | 0.989 | 0.989 | 0.989 | 0.041 | 0.997 |
| | | | Tree | 18.598 | 0.000 | **0.955** | 0.943 | 0.943 | 0.943 | 0.943 | 1.104 | 0.979 |
| 64 | 70 | 1188 | kNN | 4.211 | 7.014 | **0.999** | 0.989 | 0.989 | 0.989 | 0.989 | 0.056 | 0.996 |
| | | | Logistic Regression | 49.809 | 1.167 | 1.000 | 0.995 | 0.995 | 0.995 | 0.995 | 0.022 | 0.999 |
| | | | Neural Network | 39.913 | 2.632 | 0.999 | 0.994 | 0.994 | 0.994 | 0.994 | 0.032 | 0.999 |
| | | | Random Forest | 4.090 | 1.757 | 0.998 | 0.984 | 0.983 | 0.984 | 0.984 | 0.099 | 0.991 |
| | | | SVM | 17.891 | 4.331 | 0.999 | 0.993 | 0.993 | 0.993 | 0.993 | 0.033 | 0.995 |
| | | | Tree | 12.234 | 0.007 | 0.968 | 0.961 | 0.959 | 0.960 | 0.961 | 0.729 | 0.978 |